\documentclass{article}

\usepackage[preprint]{nips}

\usepackage{bm}

\usepackage{CJKutf8}
\usepackage{makecell}
\usepackage{graphicx}
\usepackage{amsmath}
\usepackage{amssymb}
\usepackage{algorithm, algpseudocode}
\usepackage{hyperref}
\usepackage{multirow}
\usepackage{booktabs}
\algrenewcommand\algorithmicrequire{\textbf{Input:}}
\algrenewcommand\algorithmicensure{\textbf{Output:}}
\usepackage{multicol}
\usepackage{xcolor}
\usepackage{tcolorbox}
\definecolor{rq}{HTML}{1B365C}
\definecolor{rqBack}{HTML}{9ECBF7}

\usepackage{times}
\usepackage{latexsym}

\usepackage[T1]{fontenc}
\usepackage[utf8]{inputenc}

\usepackage{microtype}

\usepackage{inconsolata}

\usepackage{graphicx}
\usepackage{tikz}
\usetikzlibrary{positioning,arrows.meta, calc,shapes.geometric}

\title{Proactive Dialogue Model with Intent Prediction}

\author{%
 Yang Luo \\
  Department of Computer Science\\
  University of Hong Kong\\
}

\begin{document}

\maketitle

\begin{abstract}
Dialogue models are inherently reactive, responding to the current user turn without anticipating upcoming intents, which leads to redundant interactions in multi-intent settings. We address this limitation by introducing a lightweight intent-transition prior derived from dialogue data and injected into the system prompt at inference time. We instantiate this prior using a Temporal Bayesian Network (T-BN) trained on per-turn intent annotations in MultiWOZ 2.2. The T-BN achieves Recall@5 $= 0.787$ and MRR $= 0.576$ on $1{,}071$ held-out USER-turn pairs. In a ground-truth replay over $200$ dialogues, BN-guided generation improves Coverage AUC from $0.742$ to $0.856$ and reduces the number of turns required to reach $75\%$ intent coverage from $3.95$ to $2.73$. These results show that lightweight intent-transition guidance enables more proactive and efficient dialogue behavior without modifying the underlying language model.
\end{abstract}

\section{Introduction}

Unlike chit-chat dialogue models that emphasize emotional support and persona consistency \cite{hong-etal-2025-dialogue}, task-oriented dialogue (TOD) systems rely fundamentally on the ability to identify and act upon user intents. While large language models (LLMs) have significantly improved response generation through strong natural language understanding and domain knowledge \cite{Brown2020LanguageModels,hong-etal-2025-qualbench}, they remain largely reactive, focusing on the current utterance without explicitly modeling how user intents evolve across turns. In practice, real-world conversations are inherently multi-intent and sequential \cite{10.1145/3774904.3792822}: users progressively reveal goals such as searching, booking, and follow-up actions. This pattern appears consistently across customer service, education \cite{10.1145/3772363.3798900, jin2025exploringimpactllmpoweredteachable}, and many task-driven scenarios. This mismatch highlights a core limitation not in generation itself, but in how intent structure is discovered, represented, and utilized.

A growing body of work has explored intent discovery and mining from dialogue data, aiming to construct meaningful intent vocabularies and uncover latent structure in user behavior \cite{10.1145/2684822.2685302,jiang2013panorama}. Early approaches rely on clustering and representation learning to group semantically similar utterances, reducing annotation costs while improving coverage of user needs \cite{lin2020discovering,hong2024neuralbayesianprogramlearningfewshot}. More recent advances emphasize large-scale mining from real interaction logs, where intent categories are not predefined but emerge from data through iterative abstraction and validation with the support of LLM utilities \cite{doi:10.36227/techrxiv.174495034.42657551/v2}, demonstrating that intent discovery can be framed as a data-driven process that combines clustering, semantic evaluation, and label refinement to construct high-quality intent taxonomies \cite{hong-etal-2025-dial}. This line of research highlights that intent is not static but context-dependent and temporally structured.

Despite progress in intent discovery, existing dialogue systems rarely leverage mined intent structures beyond per-turn classification or retrieval. Most systems treat intents as isolated labels rather than components of a dynamic process \cite{casanueva-etal-2020-efficient,Weld2022ASurvey}, ignoring how one intent leads to another. As a result, even when intent vocabularies are well-defined, systems fail to anticipate upcoming user needs and cannot exploit the sequential regularities present in dialogue data.

In this work, we improve dialogue efficiency by incorporating a data-driven prior over intent transitions into the generation process. Rather than treating intents independently, we model short-horizon dependencies between successive intents and expose this structure as an auxiliary conditioning signal during inference. Concretely, we instantiate this prior using a Temporal Bayesian Network (T-BN) trained on MultiWOZ 2.2, which estimates $P(\mathbf{x}_{t+1} \mid \mathbf{x}_t)$. At runtime, the current utterance is mapped to relevant intent nodes, and the resulting distribution over next intents is formatted as an intent-prior block and injected into the system prompt. This design biases generation toward likely follow-up intents observed in the data, without modifying the underlying LLM.

\section{Related Work}
\label{sec:related}

\paragraph{Task-oriented Dialogue System}
The MultiWOZ corpus series \cite{zang2020multiwoz22} has driven a decade of research on multi-domain task-oriented dialogue (TOD), spanning neural dialogue state trackers such as TRADE \cite{wu2019trade} and TripPy \cite{heck2020trippy}, as well as end-to-end sequence-to-sequence systems including SimpleTOD \cite{hosseini2020simpletod}, SOLOIST \cite{peng2021soloist}, and PPTOD \cite{su2022pptod}. More recently, \cite{10.1145/3774904.3792822} proposed an orchestration-free framework for customer-service automation based on flowchart-guided distillation, achieving state-of-the-art performance with lightweight language models suitable for local deployment. This line of work highlights the importance of \textbf{incorporating explicit procedural knowledge into task-oriented dialogue systems}, particularly for enforcing workflow constraints. Furthermore, speech-enabled systems have emerged, capturing user intents through ASR technologies \cite{11460729, wei-etal-2025-asr, pmlr-v260-hong25a,  lu2025contextualizedtokendiscriminationspeech}. These advances enable TOD systems to move beyond text-only interaction toward more realistic spoken dialogue scenarios.

\paragraph{Structure learning and temporal causal discovery.}
NOTEARS \cite{zheng2018notears} recasts DAG structure learning as a continuous optimization problem, enabling gradient-based induction of causal graphs. The framework has since been extended to non-parametric functional forms \cite{zheng2020notearsmlp}, graph-neural parameterizations \cite{Yu2019DAGGNNDS}, and score-based acyclicity regularizers \cite{lachapelle2020grandag}. Temporal and dynamic Bayesian networks were formalized decades earlier \cite{friedman1998dbn,murphy2002dbn}, and recent time-series extensions of NOTEARS, such as DYNOTEARS \cite{pamfil2020dynotears}, jointly learn intra- and inter-slice edges.

\paragraph{Intent discovery and proactive dialogue.}
Considering the diversity of customer expressions in dialogue interactions \cite{hong-etal-2025-augmenting}, mining true user intent is both challenging and important. Discovering the intent vocabulary has been studied through unsupervised clustering of utterance embeddings \cite{Willard2023EfficientGG, zhang2022dkn} and, more recently, through LLM-in-the-loop cluster evaluation and naming \cite{hong-etal-2025-dial, viswanathan2024large}, which support the iterative discovery of high-quality intents. Proactive dialogue systems aim to anticipate user needs rather than merely react \cite{liao2023proactive,deng2023proactivesurvey}, but most existing approaches rely on handcrafted recommendation rules or retrieval over historical trajectories.

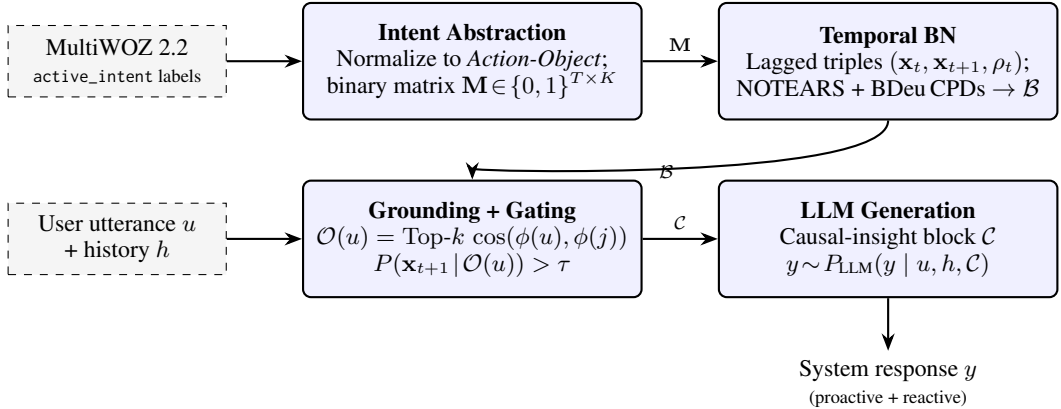
\begin{figure*}[t]
\centering
\begin{tikzpicture}[
    font=\small,
    node distance=7mm and 10mm,
    stage/.style={draw, rounded corners=3pt, fill=blue!6, thick,
                  minimum width=4.5cm, minimum height=1.55cm,
                  align=center, inner sep=4pt},
    data/.style={draw, dashed, fill=gray!8,
                 minimum width=2.9cm, minimum height=0.95cm,
                 align=center, inner sep=3pt},
    out/.style={draw, rounded corners=3pt, fill=green!8, thick,
                minimum width=2.9cm, minimum height=0.95cm,
                align=center, inner sep=3pt},
    arrow/.style={-Stealth, thick}
]

% Row 1: offline induction pipeline
\node[data] (corpus) {MultiWOZ 2.2\\
    {\scriptsize \texttt{active\_intent} labels}};
\node[stage, right=of corpus] (s1) {\textbf{Intent Abstraction}\\
    Normalize to \textit{Action-Object};\\
    binary matrix $\mathbf{M}\!\in\!\{0,1\}^{T\times K}$};
\node[stage, right=of s1] (s2) {\textbf{Temporal BN}\\
    Lagged triples $(\mathbf{x}_t,\mathbf{x}_{t+1},\rho_t)$;\\
    NOTEARS + BDeu\,CPDs $\rightarrow\mathcal{B}$};

% Row 2: runtime inference
\node[data, below=14mm of corpus] (user) {User utterance $u$\\
    + history $h$};
\node[stage, right=of user, align=center] (ground) {\textbf{Grounding + Gating}\\
$\mathcal{O}(u)=\mathrm{Top\text{-}}k\,\cos(\phi(u),\phi(j))$\\
$P(\mathbf{x}_{t+1}\!\mid\!\mathcal{O}(u))>\tau$
};

\node[stage, right=of ground, align=center] (llm) {\textbf{LLM Generation}\\
Causal-insight block $\mathcal{C}$\\
$y\!\sim\!P_{\text{LLM}}(y\mid u,h,\mathcal{C})$
};

\node[below=of llm, align=center] (resp) {System response $y$\\
{\scriptsize (proactive + reactive)}};

\draw[arrow] (corpus) -- (s1);
\draw[arrow] (s1) -- node[above,font=\scriptsize]{$\mathbf{M}$} (s2);
\draw[arrow] (user) -- (ground);
\draw[arrow] (ground) -- node[above,font=\scriptsize]{$\mathcal{C}$} (llm);
\draw[arrow] (llm) -- (resp);
\draw[arrow] (s2.south) .. controls +(0,-8mm) and +(0,1mm) ..
    node[right,font=\scriptsize,pos=0.55]{$\mathcal{B}$} (ground.north);

\end{tikzpicture}
\caption{Overview of the proposed pipeline. \textbf{Stage 1} abstracts a corpus of MultiWOZ dialogues into a binary turn--intent matrix $\mathbf{M}$. \textbf{Stage 2} lifts $\mathbf{M}$ into consecutive USER-turn pairs and fits a Temporal Bayesian Network $\mathcal{B}$ with NOTEARS under a forward-only tabu-edge constraint. At runtime, each user utterance $u$ is grounded in $\mathcal{B}$ via top-$k$ embedding similarity; the posterior $P(\mathbf{x}_{t+1}\mid\mathcal{O}(u))$ is thresholded, formatted as a causal-insight block $\mathcal{C}$, and injected into the LLM prompt so that the generated response $y$ is both reactive to the current request and anticipatory of plausible next intents.}
\label{fig:overview}
\end{figure*}

\section{Proposed Method}
\label{sec:method}

To capture probabilistic transitions between user intents, we build the model in two stages (see Figure \ref{fig:overview}): first, a static intent abstraction over the corpus, then a temporal Bayesian Network that lifts this abstraction into one-step transitions used to guide generation.

\subsection{Probabilistic Bayesian Network Abstractions}
We construct a Bayesian Network (BN) over dialogue intents in two steps: extracting an intent vocabulary and matrix from the corpus, then learning structure and parameters from this matrix.

\subsubsection{Intent Discovery}
\label{ssec:intent}

In this work, we use the ground-truth intent annotations from MultiWOZ 2.2 to demonstrate the effectiveness of BN-guided intent prediction in task-oriented dialogue systems.

\textbf{Normalization and Vocabulary}: For each USER turn, we extract the \texttt{active\_intent} from the dialogue-state \texttt{frames}. We exclude \texttt{NONE} labels and normalize surface forms into a canonical \textit{Action-Object} hyphenated format (e.g., \textit{find-restaurant}, \textit{book-hotel}). This process yields a vocabulary of $K=8$ core intents: \textit{find/book} for \textit{hotel}, \textit{restaurant}, and \textit{train}, alongside \textit{find-attraction} and \textit{find-taxi}.

\textbf{Matrix Construction}: We represent the corpus as a binary turn--intent matrix $\mathbf{M} \in \{0,1\}^{T \times K}$. Each entry is defined as:
\begin{equation}
    M_{t,j} = \mathbb{I}(\text{intent}_j \text{ is active at turn } t)
\end{equation}
where $T$ is the total number of USER turns in the corpus. Each row is tagged with its parent dialogue identifier and within-dialogue turn index, which are required to form temporal pairs in Stage~2.

\subsubsection{Structure and Parameter Learning}
Given the intent matrix $M$, we learn the directed acyclic graph (DAG) structure and the underlying probability distributions:

\textbf{Structure Learning}: We use the NOTEARS (Non-combinatorial Optimization for Learned Acyclic Graphs) algorithm to learn the DAG. Edge weights $W$ below a threshold are pruned to keep the graph sparse and interpretable.

\textbf{Parameter Estimation}: Once the structure is defined, we fit the Conditional Probability Distributions (CPDs) for each node using a \textbf{Bayesian Estimator}. This allows the model to estimate the conditional probability $P(I_i \mid Pa(I_i))$, where $Pa(I_i)$ denotes the parent nodes of intent $I_i$, capturing the probabilistic nature of user behavior.

\subsection{Knowledge Injection}
\label{ssec:tbn}

Based on the processed intent matrix, we construct a temporal Bayesian Network (BN) to model the probabilistic transitions between user intents.

\subsubsection{Lagged Design Matrix}
To capture transitions rather than mere co-occurrence, we transform $\mathbf{M}$ into a lagged matrix representing consecutive USER-turn pairs $(t, t{+}1)$. Each instance is encoded as a tuple $\big[\mathbf{x}_t,\ \mathbf{x}_{t+1},\ \rho_t\big]$, where $\mathbf{x}_t, \mathbf{x}_{t+1} \in \{0,1\}^{K}$ are intent indicators and $\rho_t$ is a categorical \textit{progress} feature derived by bucketing the normalized turn index $t / T_{\max}$ into three bins: $\{\text{early, mid, late}\}$. For notation, we denote temporal copies of intent $j$ as $j_{t}$ and $j_{t+1}$.

\subsubsection{Structure and Parameter Learning}
We learn the DAG $\mathcal{G}$ using the NOTEARS framework over $2K + 3$ variables. To ensure physical causality, we impose a \textit{tabu-edge constraint} that forbids any back-step edges ($j_{t+1} \rightarrow i_{t}$). Following structure induction, we fit the conditional probability distributions (CPDs) using a Bayesian Estimator with a BDeu prior. This network $\mathcal{B}$ allows us to compute the one-step transition distribution:
\begin{equation}
P(\mathbf{x}_{t+1} \mid \mathbf{x}_t, \rho_t)
\end{equation}
which serves as the probabilistic prior for the subsequent generation stage.

\subsubsection{BN-Guided Proactive Generation}
During dialogue execution, the \texttt{TemporalBNReasoner} generates proactive guidance through the following four-step pipeline:

\textbf{(a) Utterance Grounding}: The user utterance $u$ is embedded via $\phi(u)$ to compute cosine similarity against canonical intent names. The top-$k$ nodes ($k=5$) form the observation set $\mathcal{O}(u) = \text{Top-}k\{\cos(\phi(u), \phi(j))\}_{j \in \mathcal{V}_t}$.

\textbf{(b) Intent Inference}: For each $j_t \in \mathcal{O}(u)$, we set evidence $j_t=1$ and perform variable elimination to derive the posterior distribution over future intents $P(\mathbf{x}_{t+1} \mid \mathcal{O}(u))$.

\textbf{(c) Probabilistic Gating}: We filter predictions using a confidence threshold (e.g., $P > 0.5$). Only high-probability intents are retrieved as ``proactive insights,'' preventing the LLM from volunteering low-confidence or irrelevant suggestions.

\textbf{(d) Causal-insight Injection}: The surviving insights are formatted as a \textit{causal-insight} block and prepended to the system prompt. The final response $y$ is sampled as:
\begin{equation}
y \sim P_{\text{LLM}}(y \mid u, h, \mathcal{C})
\end{equation}
where $h$ is the dialogue history and $\mathcal{C}$ is the BN-derived guidance. Two heuristics keep the prompt clean: if the gate returns nothing, no insight block is emitted, and the user's immediate request always takes precedence over any anticipated next intent.

\section{Experimental Setup}

\subsection{Benchmark Dataset}
\label{ssec:dataset}

We conduct all experiments on MultiWOZ 2.2, a large multi-domain task-oriented dialogue corpus spanning seven domains (\textit{hotel}, \textit{restaurant}, \textit{train}, \textit{taxi}, \textit{attraction}, \textit{police}, \textit{hospital}). Intent labels are taken directly from the per-turn \texttt{active\_intent} annotation.

We partition the corpus at the dialogue level under a fixed random seed (seed $= 42$) so that no dialogue contributes simultaneously to structure learning and to evaluation. Two disjoint held-out subsets are drawn from the test split: (i) a \emph{knowledge-abstraction} subset of $200$ dialogues, yielding $1{,}071$ consecutive USER-turn pairs $(\mathbf{x}_t, \mathbf{x}_{t+1})$ across the $K{=}8$ canonical intents, on which we evaluate the intrinsic predictive fidelity of the induced temporal BN under a deterministic 80/20 dialogue-level split; and (ii) the same $200$ held-out dialogues are used for a \emph{ground-truth replay} Track B evaluation in which BN next-intent predictions are compared against the actual subsequent intents without any LLM generation call, enabling a large-scale and reproducible efficiency assessment. The remaining training dialogues are used exclusively for Stage-2 structure and parameter learning.

\subsection{Evaluation Metrics}
\label{ssec:metrics}

We evaluate along two complementary axes: the intrinsic quality of the learned temporal abstraction (Track~A), and the extrinsic quality of dialogues produced when the abstraction is used to steer an LLM (Track~B).

\paragraph{Track A -- Knowledge Abstraction.} 
We evaluate the intrinsic predictive fidelity of the induced temporal BN using standard ranking metrics and structural diagnostics. 
\textbf{Ranking Performance} is measured by Recall@$k$ ($k \in \{1,3,5\}$) and Mean Reciprocal Rank (MRR) of the ground-truth next intent. 
\textbf{Structural Soundness} is assessed via edge stability under 5-fold cross-validation and the count of backward edges (\texttt{\_\_t1} $\rightarrow$ \texttt{\_\_t}) to verify causal constraints.

\paragraph{Track B -- End-to-end Dialogue Quality.} 
We assess the downstream impact through both human-centric and task-oriented dimensions.
\textbf{Subjective Quality} is rated on a 1--5 Likert scale for \textit{Anticipation} and \textit{Helpfulness} by a blinded LLM-judge.
\textbf{Objective Performance} is quantified by Intent Hit Rate, Jaccard Similarity, and dialogue efficiency metrics, specifically Average Turns ($T$) and Turn-Reduction Rate ($\Delta_{\text{turn}}$).

\subsection{Baselines}
To isolate the contribution of the BN guidance signal, all systems share the same LLM backbone (\texttt{MiniMax-M2.5}).  
We compare our \textbf{Temporal BN} against:
(i) \textbf{Plain-LLM}, a context-conditioned generator without BN guidance;
(ii) \textbf{Random Transition}, which draws next-intents uniformly from the vocabulary;
(iii) \textbf{Marginal (no-lag)}, which relies on empirical frequencies $P(\mathbf{x}_{t+1})$ without temporal conditioning;
and (iv) \textbf{Bigram (lag-1 context)}, which conditions on both the previous intent $\mathbf{x}_{t-1}$ and the current intent $\mathbf{x}_t$, with backoff to marginal when the bigram context is unseen.

\section{Results and Discussion}
\label{sec:results}

\subsection{Main Results}
\label{ssec:main-results}

We evaluate the temporal BN as an inference-time prior over $200$ held-out dialogues via \emph{ground-truth replay}: the BN's top-$k$ predicted next intents are compared directly against the actual subsequent USER-turn intents from the MultiWOZ test split, with no LLM generation in the loop. Removing the generator removes LLM-judge variance and makes the comparison reproducible at scale.

\begin{table}[t]
\centering
\small
\caption{Main results on MultiWOZ 2.2 under ground-truth replay over $n{=}200$ held-out dialogues (threshold $\tau{=}0.5$, top-$k{=}5$). Baseline = cumulative coverage using only actual user intents; BN-Guided = coverage augmented by BN top-5 predictions.}
\label{tab:main}
\begin{tabular}{lccc}
\toprule
\textbf{Metric} & \textbf{Baseline} & \textbf{BN-Guided} & \textbf{$\Delta$} \\
\midrule
Coverage AUC ($\uparrow$)             & 0.742 & \textbf{0.856} & $+0.114$ \\
Turns to 75\% coverage ($\downarrow$) & 3.95  & \textbf{2.73}  & $-1.22$  \\
Turn Reduction ($\uparrow$)           &---   & $+1.04 \pm 1.69$ & ---    \\
Intent Hit Rate ($\uparrow$)          & ---   & $0.538 \pm 0.235$ & ---   \\
Intent Jaccard ($\uparrow$)           & ---   & $0.185 \pm 0.092$ & ---   \\
\bottomrule
\end{tabular}
\end{table}

The BN-guided arm raises Coverage AUC by $+0.114$ (from $0.742$ to $0.856$) and shortens the path to $75\%$ intent coverage by $1.22$ turns ($3.95 \to 2.73$, a $-30.9\%$ relative reduction). The per-dialogue turn reduction is $+1.04 \pm 1.69$, with most of the gain concentrated in multi-intent dialogues where proactive anticipation can resolve a later goal before the user explicitly states it.

\subsection{Performance of Knowledge Abstraction}
\label{ssec:ka-results}

We now evaluate the temporal BN \emph{intrinsically}, independent of any downstream LLM. Table~\ref{tab:kb} reports next-intent prediction on the $1{,}071$ held-out USER-turn pairs described in Section~\ref{ssec:dataset}.

\begin{table}[t]
    \centering
    \small
    \caption{Next-intent prediction on $1{,}071$ held-out USER-turn pairs from $200$ dialogues (threshold $\tau{=}0.5$). Recall@$k$ measures whether the ground-truth next intent appears in the top-$k$ predictions of each method. The Bigram baseline conditions on both the previous and current intent.}
    \label{tab:kb}
    \begin{tabular}{lcccc}
    \toprule
    \textbf{Method} & \textbf{R@1} & \textbf{R@3} & \textbf{R@5} & \textbf{MRR} \\
    \midrule
    Random Transition          & $0.107$ & $0.325$ & $0.517$ & $0.324$ \\
    Marginal (no-lag)          & $0.220$ & $0.518$ & $0.783$ & $0.461$ \\
    Bigram (lag-1 context)     & $0.167$ & $0.421$ & $0.583$ & $0.357$ \\
    \textbf{Ours (Temporal BN)} & $\mathbf{0.396}$ & $\mathbf{0.595}$ & $\mathbf{0.787}$ & $\mathbf{0.576}$ \\
    \bottomrule
    \end{tabular}
\end{table}

The Temporal BN reaches Recall@5 $= \mathbf{0.787}$ and MRR $= \mathbf{0.576}$, above the random baseline ($0.517$ / $0.324$) and the marginal baseline ($0.783$ / $0.461$). The MRR gap ($0.576$ vs.\ $0.461$) matters because the Stage-3 guidance prompt only consumes the top-ranked intents, so placing the true next intent near the top --- not merely inside the top-5 --- is what determines whether the prompt actually helps. Conditioning on $\mathbf{x}_t$ therefore yields a $+24.9\%$ relative MRR gain over the lag-free baseline that models only $P(\mathbf{x}_{t+1})$. The Bigram baseline (R@5 $= 0.583$) underperforms Marginal (R@5 $= 0.783$): many $(I_{t-1}, I_t)$ contexts appear only once in training, so its backoff fires too often. The BN's structured graph supplies more stable conditioning than raw n-gram counts.

\paragraph{Structural soundness.} The tabu-edge constraint introduced in Section~\ref{ssec:tbn} drives the number of backward (\texttt{\_\_t1}$\rightarrow$\texttt{\_\_t}) edges in the learned graph to \textbf{zero}, confirming that forward-only causality is hard-enforced rather than softly preferred. Under 5-fold cross-validation, $\mathbf{6/8}$ of the temporal self-persistence edges are robust in the sense of appearing in $\geq 4/5$ folds with a weight standard deviation below $0.10$, and all robust edges have $100\%$ cross-fold consistency. The strongest self-persistence edge is \texttt{find-taxi\_\_t}$\rightarrow$\texttt{find-taxi\_\_t1} ($w = 0.59 \pm 0.06$), reflecting the tendency of users to iteratively refine taxi queries across consecutive turns.

\paragraph{Per-intent breakdown.} Recall@5 is not uniform across the $8$ intents. \texttt{find-restaurant}, \texttt{find-train}, \texttt{find-attraction} and \texttt{find-hotel} achieve a perfect $1.00$; \texttt{book-restaurant} reaches $0.89$; but \texttt{find-taxi} ($0.57$), \texttt{book-hotel} ($0.36$), and especially \texttt{book-train} ($0.00$) lag behind. The two failing \texttt{book-*} intents coincide with the rarest transitions in the training corpus, and we attribute the gap to data sparsity: with only $K{=}8$ intents and a large imbalance in their marginal frequencies, NOTEARS is under-regularised for tail transitions. Addressing this will require either a larger training slice of MultiWOZ or an explicit re-balancing of rare transitions in the lagged matrix, both of which we leave to future work.

\paragraph{Structural quality.}
At threshold $\tau{=}0.5$, the learned graph contains $29$ edges across $22$ nodes (average degree $2.64$, within the interpretable $[2,4]$ target band). Two of the eight temporal self-persistence edges exceed the $0.2$-bit information gain threshold (\texttt{find-train}: $0.265$ bits; \texttt{find-restaurant}: $0.245$ bits). Over $20$ random-DAG trials with the same node/edge counts, the learned BN achieves Recall@5 $= 0.787$ versus $0.550 \pm 0.312$ for random structures ($+43.1\%$), confirming that the learned causal graph is substantially more predictive than a random baseline.

The Track-A results support two claims. First, a temporal BN induced from ground-truth intent annotations under a tabu-edge constraint gives a structurally clean and empirically strong model of short-horizon intent dynamics in multi-domain dialogue. Second, the gain at the BN level is larger than what propagates into end-to-end dialogue quality (Section~\ref{ssec:main-results}), pointing to the probabilistic-guidance-to-LLM interface --- not the abstraction itself --- as the next thing to improve.

\section{Conclusion}

We present a simple yet effective framework that bridges intent structure modeling and dialogue generation through a Temporal Bayesian Network (T-BN). By explicitly capturing short-horizon intent transitions and injecting them as probabilistic guidance at inference time, the proposed approach enables more proactive and efficient task-oriented dialogue without modifying the underlying LLM. Empirical results on MultiWOZ 2.2 show that the T-BN achieves strong intrinsic predictive performance and consistently improves dialogue efficiency in ground-truth replay. These findings suggest that lightweight, interpretable intent-transition models can complement large language models by providing structured foresight. Future work will focus on improving robustness for rare intent transitions, extending to longer temporal dependencies, and refining the interface between probabilistic guidance and generation to better translate modeling gains into end-to-end dialogue quality.

\bibliographystyle{unsrt}
\bibliography{custom}

\end{document}